\documentclass{article}

\usepackage{arxiv}

\usepackage[utf8]{inputenc} 
\usepackage[T1]{fontenc}    
\usepackage{hyperref}       
\usepackage{url}            
\usepackage{booktabs}       
\usepackage{amsfonts}       
\usepackage{nicefrac}       
\usepackage{microtype}      
\usepackage{lipsum}		

\usepackage{graphicx}
\usepackage{subfig} 
\usepackage{svg}
\usepackage{amsmath}

\newcommand{\X}{\mathcal{X}}
\newcommand{\Y}{\mathcal{Y}}
\renewcommand{\L}{\mathcal{L}}

\newcommand{\Pro}{\mathrm{P}}
\newcommand{\E}{\mathrm{E}}

\usepackage{booktabs,siunitx}

\clearpage{\thispagestyle{empty}\cleardoublepage}
\usepackage{adjustbox}
\usepackage{longtable}
\usepackage{float}
\usepackage{graphicx}
\usepackage{flafter}

\usepackage{listings}
\usepackage{color}
\title{Unsupervised Latent Space Translation Network}

\date{} 					

\author{
  Magda Friedjungová \\
  Faculty of Information Technology\\
  Czech Technical University in Prague\\
  Prague, Czech Republic\\
  \texttt{magda.friedjungova@fit.cvut.cz} \\
   \And
  Daniel Vašata \\
  Faculty of Information Technology\\
  Czech Technical University in Prague\\
  Prague, Czech Republic\\
  \texttt{daniel.vasata@fit.cvut.cz} \\
  \And
  Tomáš Chobola \\
  Faculty of Information Technology\\
  Czech Technical University in Prague\\
  Prague, Czech Republic\\
  \texttt{choboto1@fit.cvut.cz} \\
  \And
  Marcel Jiřina \\
  Faculty of Information Technology\\
  Czech Technical University in Prague\\
  Prague, Czech Republic\\
  \texttt{marcel.jirina@fit.cvut.cz} \\
}

\date{}

\renewcommand{\undertitle}{A preprint of paper accepted to ESANN 2020.}

\begin{document}
\maketitle


\begin{abstract}
One task that is often discussed in a computer vision is the mapping of an image from one domain to a corresponding image in another domain known as image-to-image translation. Currently there are several approaches solving this task. In this paper, we present an enhancement of the UNIT framework that aids in removing its main drawbacks. More specifically, we introduce an additional adversarial discriminator on the latent representation used instead of VAE, which enforces the latent space distributions of both domains to be similar. On MNIST and USPS domain adaptation tasks, this approach greatly outperforms competing approaches.
\end{abstract}

\section{Introduction}
The problem of mapping images between different domains can be tackled in both a supervised and an unsupervised manner. In the supervised approach, one needs pairs of corresponding images in both domains to learn the model. In real-world datasets, these pairs have to be created somehow, which leads to a very challenging problem.
On the other hand, the unsupervised approach is used when working with independent unpaired sets of images. The difficulty lies in the fact that there are no paired examples demonstrating how the images should be mapped to each other. Hence,
the correctness of an image mapped from one domain to another is usually estimated using an implicitly learned probability distribution in the second domain.

From a general point of view the problem is a part of transfer learning \cite{Pan2009} which
focuses on storing knowledge gained when solving one problem and applying it to a different but related problem.
Transfer learning methods are often used in image processing where one suffers from lack of labeled data, computational difficulties, differences in data representation, color settings, etc.

We consider a scenario where two image domains differ in feature representations but the target
supervised prediction task is the same. Specifically, we assume that both domains are rich with data but only one is equipped with labels. Transfer learning approaches enable us to modify the domain without labels in such a manner that allows it to be represented in the same way as the domain with labels, taking advantage of an already trained prediction model.

In our research we propose a novel Latent Space Translation Network (LSTNet) based on shared latent space representation and adversarial training inspired by the Unsupervised Image-to-image Translation (UNIT) framework \cite{UNIT}. However, we do not use a variational autoencoder (VAE)
as a component and instead introduce another adversarial discriminator which attempts to guess from a latent space representation
of an image which domain it is from. This approach enforces the encoders from source domains to latent space representations
to yield the same distribution for both domains. For this to work one needs a shared latent space assumption,
which means that a pair of corresponding images
in the two domains can be mapped to the same latent representation in a shared-latent space, see also \cite{COGAN}.

\subsection{Related Work}
Many recent works \cite{COGAN,PixelDA,UNIT,CYCADA} are primarily focused on unsupervised domain adaptation in image processing using GANs \cite{GAN}. In domain adaptation, rich labeled data are leveraged on a source domain to achieve performance on a target domain regardless of unlabeled or poorly labeled data. 

The architecture called CoGAN \cite{COGAN} applies GANs to the domain transfer problem by training two coupled GANs to generate the source and target images, respectively. The approach achieves a domain invariant feature space by tying the high-level layer parameters of the two GANs learned a joint distribution without any tuple of corresponding images with just samples drawn from the marginal distributions and shows that the same noise input can generate a corresponding pair of images from these two distributions.

Recently it was shown that generative adversarial networks combined with cycle-consistency constraints \cite{Zhu2017} are very effective in mapping data between different domains, even without the use of aligned data pairs.
A very successful model in particular is the UNIT framework \cite{UNIT}.
Each image domain is modeled using a VAE-GAN. The adversarial training objective interacts with a weight-sharing constraint,
which enforces a shared latent space to generate corresponding images in two domains, while the VAEs
relate translated images with input images in the respective domains.


\section{Translation Network}
Let us denote by $\X_1$ the source image domain with associated labels in some label space $\Y$. Similarly, let $\X_2$ be the target image domain, but with unknown labels. The goal of the domain adaptation is to learn the predictive function $f: \X_2 \to \Y$ in the target domain by leveraging the information from the source domain.
Therefore, we consider a source domain dataset $\{(x^{(i)}_1, y^{(i)}) \in \X_1 \times \Y\mid i=1, \ldots, n_1\}$
consisting of image-label pairs and a target domain dataset $\{x^{(j)}_{2} \in \X_2 \mid i=1,..., n_2\}$ with no labels. 
In the unsupervised approach to domain adaptation one starts by learning the mapping $g: \X_2 \to \X_1$
based on independent datasets in $\X_1$ and $\X_2$.
Then the desired prediction function is given by the composition of $g$ with a predictive function $h$ in $\X_1$, $f = h \circ g$, which can be estimated because we have labels in the source domain dataset.

Let us now focus on finding a suitable function $g$. Using the latent space assumption one can construct
such a function as the composition of the encoder function $E_2: \X_2 \to \L$, mapping images from target space $\X_2$
to shared latent space $\L$, with the generator function $G_1: \L \to \X_1$ mapping points in the shared latent space $\L$ to
source space $\X_1$. In order to be able to train these functions in an unsupervised manner it is useful to have an encoder
$E_1: \X_1 \to \L$ and a generator $G_2: \L \to \X_2$ which enables one to apply cycle consistency constraints, \cite{Zhu2017}, given by $x_1 = G_1(E_1(x_1))$, $x_2 = G_2(E_2(x_2))$, $x_1 = G_1(E_2(G_2(E_1(x_1))))$, $x_2 = G_2(E_1(G_1(E_2(x_2))))$.

Similarly to \cite{UNIT} the LSTNet itself consists of six subnetworks including two domain encoders $E_1$, $E_2$, two image generators $G_1$, $G_2$ and two domain adversarial discriminators $D_1$, $D_2$. The encoder is responsible for mapping an input image to a code in latent space $\L$, which is taken by the generator which then reconstructs the image. Discriminators are trained to differentiate between real and fake images for each domain, whereas the generators are trained to fool them.

Since we assume that there is one-to-one correspondence between images in both domains we may expect that the probability distributions $\Pro(E_1(x_1))$ and $\Pro(E_2(x_2))$ of points in the shared latent space $\L$ mapped from $\X_1$ and from $\X_2$ are similar.
To achieve this we introduce another adversarial discriminator $D_l$ trying to differentiate between points from source and target domains based on their latent space representations. This eliminates one of the drawbacks of the UNIT framework which is the Gaussian latent space assumption enforced by VAE components.

\begin{figure}[h!]
\centering
\includegraphics[scale=0.5]{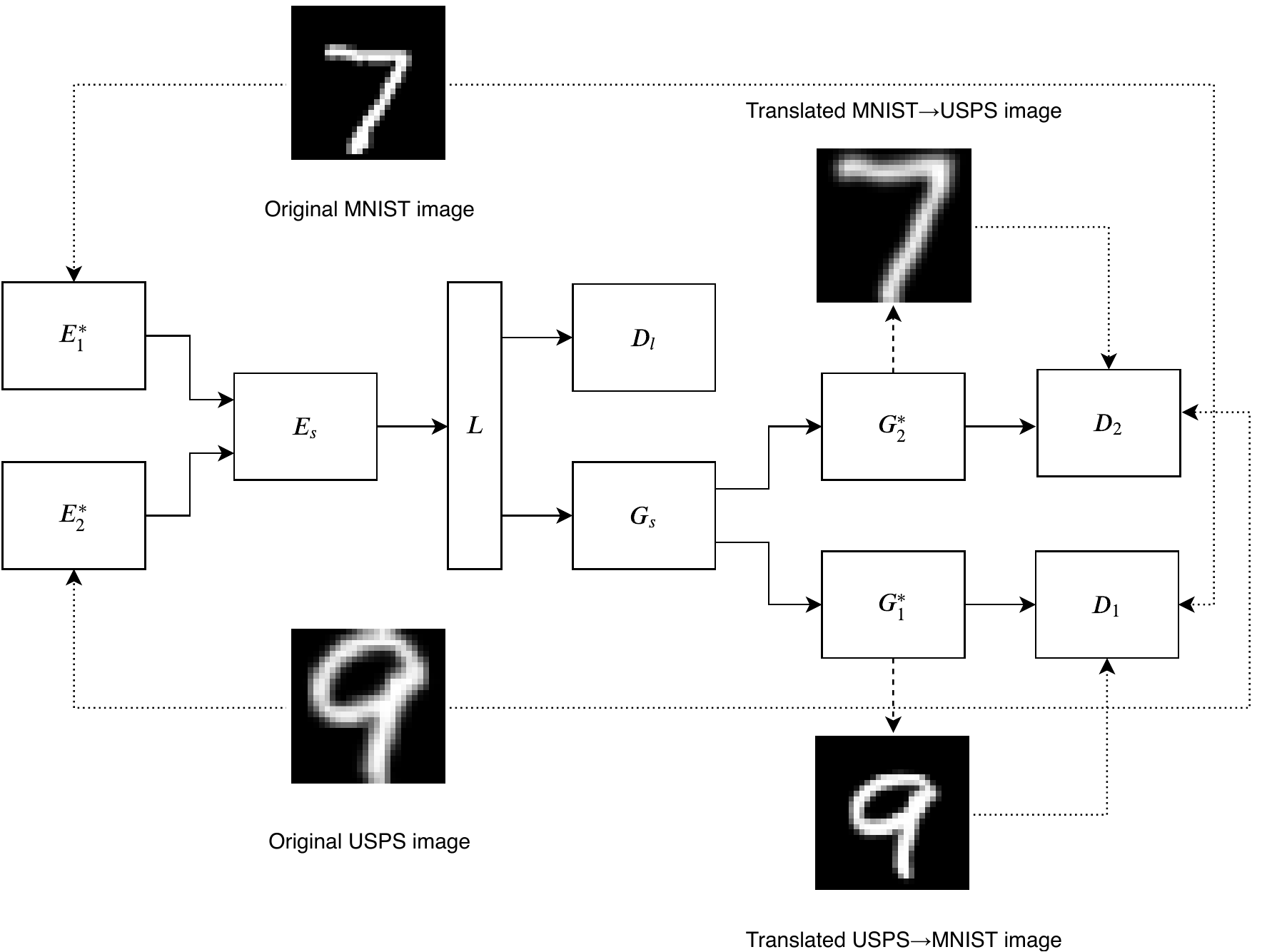}
\caption{Architecture of proposed LSTNet with MNIST-USPS example.}
\label{fig:unit}
\end{figure}

Furthermore, in order to support the latent space assumption, we assume shared intermediate representations
of both encoders $E_1, E_2$ and generators $G_1, G_2$.
Hence, we have $E_1 = E_s \circ E^*_1$ and $E_2 = E_s \circ E^*_2$, where $E_s$ is the shared component of both encoders $E_1, E_2$
and $E^*_1$ and $E^*_2$ are the custom components of $E_1$ and $E_2$, respectively. A similar composition holds for the generators,
i.e. $G_1 = G^*_1 \circ G_s$ and $G_2 = G^*_2 \circ G_s$, where $G_s$ is the shared component of both generators $G_1, G_2$,
and $G^*_1$ and $G^*_2$ are the custom components of $G_1$ and $G_2$, respectively.
A schematic depiction of the entire network is given in Figure \ref{fig:unit}.

\subsection{Training}
For the training we may identify three subnetworks: $\text{AN}_1 = (E_2, G_1, D_1)$, $\text{AN}_2 = (E_1, G_2, D_2)$,
$\text{AN}_l = (E_1, E_2, D_l)$.
$\text{AN}_1$ is responsible for distinguishing real images sampled from $\X_1$ from images sampled from $\X_2$ and translated to $\X_1$ by the mapping $G_1 \circ E_2$.
Analogously, $\text{AN}_2$ is responsible for distinguishing real images sampled from $\X_2$ from images sampled from $\X_1$ and translated to $\X_2$ by the mapping $G_2 \circ E_1$.
$\text{AN}_l$ is trying to find out which source domain the current point in the latent space corresponds to. Therefore, it should output $1$ (true) for images sampled from $\X_1$ mapped by $E_1$ into $\L$ and $2$ (false) for images sampled from $\X_2$ mapped by $E_2$.

It should be mentioned that in our case none of the three subnetworks are a proper GAN.
This is because they are not generative - we never let the generators transform random inputs into images.

The learning consists of the simultaneous optimization of objective functions corresponding to adversarial training of the three networks $\text{AN}_1$, $\text{AN}_2$, $\text{AN}_l$ and
objective functions corresponding to four cycle consistency conditions:
$\text{id}_{\X_1} = G_1 \circ E_1$, $\text{id}_{\X_2} = G_2 \circ E_2$,
$\text{id}_{\X_1} = G_1 \circ E_2 \circ G_2 \circ E_1$, and
$\text{id}_{\X_2} = G_2 \circ E_1 \circ G_1 \circ E_2$.
Hence, we want to minimize the weighted sum of particular objectives
\begin{multline}\label{eq:objective}
  J(E_1, E_2, G_1, G_2, D_1, D_2, D_l) = w_1 J_{\text{AN}_1}(E_2, G_1, D_1)
  + w_2 J_{\text{AN}_2}(E_1, G_2, D_2)\\
 + w_l J_{\text{AN}_l}(E_1, E_2, D_l)
  + w_3 J_{\text{CC}_1}(E_1, G_1) + w_4 J_{\text{CC}_2}(E_2, G_2)\\
  + w_5 J_{\text{CC}_3}(E_1, E_2, G_1, G_2)
  + w_6 J_{\text{CC}_4}(E_1, E_2, G_1, G_2),
\end{multline}
where objective functions for adversarial networks are
\[
\begin{aligned}
  J_{\text{AN}_1}(E_2, G_1, D_1) = \E_{x_1 \sim \Pro_{\X_1}} \log D_1(x_1)
  + \E_{x_2 \sim \Pro_{\X_2}} \log\big(1 - D_1(G_1(E_2(x_2)))\big),\\
  J_{\text{AN}_2}(E_1, G_2, D_2) = \E_{x_2 \sim \Pro_{\X_2}} \log D_2(x_2)
  + \E_{x_1 \sim \Pro_{\X_1}} \log\big(1 - D_2(G_2(E_1(x_1)))\big),\\
  J_{\text{AN}_l}(E_1, E_2, D_l) = \E_{x_1 \sim \Pro_{\X_1}} \log D_l(E_1(x_1))
  + \E_{x_2 \sim \Pro_{\X_2}} \log\big(1 - D_l(E_2(x_2))\big)
\end{aligned}
\]
and objective functions for cycle consistency conditions are given by MAE:
\[
\begin{aligned}
  J_{\text{CC}_1}(E_1, G_1)  &= \E_{x_1 \sim \Pro_{\X_1}}\lVert x_1 - G_1(E_1(x_1))\rVert_1,\\
  J_{\text{CC}_3}(E_1, E_2, G_1, G_2)  &= \E_{x_1 \sim \Pro_{\X_1}}\lVert x_1 - G_1(E_2(G_2(E_1(x_1))))\rVert_1,\\
\end{aligned}
\]
and analogously for $J_{\text{CC}_2}(E_2, G_2)$ and $ J_{\text{CC}_4}(E_1, E_2, G_1, G_2)$.

The training represents a two team adversarial game, where the first team consists of encoders and generators,
and the second team consists of discriminators. The optimization is done via alternating gradient descent, where the first step is updating the discriminators $D_1, D_2$, and $D_l$, and the second step is updating the encoders $E_1, E_2$ and generators $G_1, G_2$.


\section{Experiments}
We performed the experiments on benchmark datasets MNIST \cite{MNIST} and USPS \cite{USPS} devoted to digit classification, which were used in previous related studies \cite{UNIT,COGAN,DEEPJDOT}. For both domains, we used the entire training sets, i.e. 60000 training images for MNIST and 7291 for USPS. Test sets contain 10000 MNIST images and 2007 USPS images. Both datasets consist of grayscale images, the size of MNIST images is 28x28 and of USPS is 16x16.

In the first step, the LSTNet was trained using images from both domains without knowledge of labels. As an optimizer we used Adam with a learning rate of $0.0001$ and moment estimates exponential decays $0.8$ and $0.999$. Mini-batches were of size $64$ images from each domain. We also used data augmentation with randomly rotated training images by a maximum of $10$ degrees, rescaled by a random number in the range of $[0.9,1.1]$, and shifted randomly by a maximum of 2 pixels in each direction.
The weights corresponding to the objective function \eqref{eq:objective} were chosen to be
$w_1, w_2=20$, $w_l=30$, and $w_3, w_4, w_5, w_6=100$. A description of the architecture details is given in Table \ref{netarchitecture}.

\begin{table}[h!]
  \centering
  \tiny
  \begin{tabular}{|c|c|c|}
    \hline
    Layer & Encoders & Shared \\
    \hline
    1& CONV-(N64, K7, S1), BatchNorm, LeakyReLU & No \\
    2& CONV-(N128, K5, S2), BatchNorm, LeakyReLU & No \\
    3& CONV-(N256, K3, S2/S1), BatchNorm, LeakyReLU & No \\
    4& CONV-(N512, K3/K2-V, S1), BatchNorm, LeakyReLU & No \\
    5& CONV-(N256, K3, S1), BatchNorm, LeakyReLU & Yes \\
    6& CONV-(N128, K3, S3), BatchNorm, LeakyReLU & Yes \\
    \hline
    Layer & Generators & Shared \\
    \hline
    1& DCONV-(N128, K3, S1), BatchNorm, LeakyReLU & Yes \\
    2& DCONV-(N256, K3, S1), BatchNorm, LeakyReLU & Yes \\
    3& DCONV-(N512, K3/K2-V, S1), BatchNorm, LeakyReLU & No \\
    4& DCONV-(N256, K3, S2), BatchNorm, LeakyReLU & No \\
    5& DCONV-(N128, K5, S2/S1), BatchNorm, LeakyReLU & No \\
    6& DCONV-(N64, K7, S1), BatchNorm, LeakyReLU & No \\
    7& DCONV-(N1, K1, S1), TanH & No \\
    \hline
    Layer & Discriminators & Shared \\
    \hline
    1& CONV-(N64, K3, S1), LeakyReLU, MaxPooling-(K2, S1) & No \\
    2& CONV-(N128, K3, S1), LeakyReLU, MaxPooling-(K2, S2/S1)  & No \\
    3& CONV-(N256, K5, S1), LeakyReLU, MaxPooling-(K2, S2)  & No \\
    4& CONV-(N512, K3/K2-V, S1), LeakyReLU, MaxPooling-(K2, S2)  & No \\
    5& FC-(N1), Sigmoid & No \\
    \hline
    Layer & Latent Discriminator & Shared \\
    \hline
    1& CONV-(N256, K3, S1), LeakyReLU, MaxPooling-(K2, S1)  & No \\
    2& CONV-(N512, K3, S1), LeakyReLU, MaxPooling-(K2, S2)  & No \\
    3& CONV-(N256, K3, S1), LeakyReLU, MaxPooling-(K2, S1)  & No \\
    4& FC-(N1), Sigmoid & No \\
    \hline
  \end{tabular}
  \caption{Architecture details of the translation network. Abbreviation: DCONV=transposed convolutional layer, FC=fully connected layer, N=neurons, K=kernel size, S=stride size, V= "valid" padding instead of default "same" padding. Slash is used to distinguish the first and second domain. }
  \label{netarchitecture}
\end{table}

In the second step, the classification model was trained on the MNIST training dataset in a supervised manner (accuracy achieved on a test set was 0.9941). Then the USPS test dataset was translated into the MNIST domain using a previously trained translation network. The classification model was tested on this translated dataset and achieved an accuracy of 0.9701. Similarly, we trained a classifier on USPS (accuracy 0.9751) and then evaluated it on the translated MNIST test dataset (accuracy 0.9761).
The comparison of our results and results presented in \cite{COGAN,UNIT,DEEPJDOT} is given in Table \ref{results}. We achieved significantly better results in both the USPS to MNIST and the MNIST to USPS translations.
\begin{table}[h!]
  \centering
  \tiny
  \begin{tabular}{|c|c|c|c|c|}
    \hline
    Method & CoGAN~\cite{COGAN} & UNIT~\cite{UNIT} & DeepJDOT~\cite{DEEPJDOT} &  Proposed LSTNet \\
    \hline
    USPS $\to$ MNIST & 0.9315 & 0.9358 & 0.964 & \textbf{0.9701}\\
    MNIST $\to$ USPS & 0.9565 & 0.9597 & 0.957 & \textbf{0.9761}\\
    \hline
  \end{tabular}
  \caption{Comparison of accuracies of methods used in unsupervised domain adaptation.}
  \label{results}
\end{table}


\section{Conclusion}
We propose LSTNet as a novel framework based on shared latent space representation and adversarial training.
Our work is inspired by the UNIT framework. However, in contrary to UNIT, instead of using VAEs we introduce an additional adversarial discriminator on the latent representation which forces the latent space distributions from both domains to be similar. We experimentally showed an interesting performance enhancement of the proposed network in the domain adaptation of MNIST and USPS datasets. In future work we would like to focus on the use of LSTNet on other domain adaptation tasks.

\section*{Acknowledgements}
This research has been supported by SGS grant No. SGS17/210/OHK3/3T/18 and by GACR grant No. GA18-18080S.


\bibliography{arxiv}{}
\bibliographystyle{unsrt}

\end{document}